%% file: template.tex
\documentclass[cameraready]{Interspeech}


\usepackage{comment}
\usepackage{amsmath}
\usepackage{amssymb}
\usepackage{booktabs}
\usepackage{multirow}
\usepackage{makecell}
\usepackage{graphicx}
\usepackage{enumitem}
\usepackage{xcolor}
\usepackage{tikz}
\usepackage{pgfplots}
\usetikzlibrary{plotmarks} 
\pgfplotsset{compat=1.16}
\usetikzlibrary{decorations.pathreplacing, positioning, arrows.meta, shapes.geometric, fit, backgrounds, calc}

\definecolor{verbatifmRed}{HTML}{D9534F}   
\definecolor{intendedBlue}{HTML}{428BCA}  
\definecolor{connectorGray}{HTML}{B0B0B0}

\title{Transcription Policy as a Latent Variable:\\ Activating Controllable Verbatim ASR with Word-Level Timing}

\author[affiliation={1}, correspondingauthor]{Laurin}{Wagner}
\author[affiliation={1}]{Mario}{Zusag}
\author[affiliation={1}]{Bernhard}{Thallinger}

\address{
    $^1$ nyra labs, Austria
}

\email{lwagner@nyra-labs.com, mzusag@nyra-labs.com, bthallinger@nyra-labs.com}


\begin{document}
\maketitle

\begin{abstract}
Modern ASR models trained on heterogeneously annotated data treat transcription style (verbatim vs. intended) as an uncontrolled latent variable, causing measurable decoding instability, evaluation confounding (up to 60\% of reported WER from style mismatch), and unreliable word-level timing. We show models already encode both styles, the challenge is controlled activation. Using coverage-aware decoder task tokens trained on parallel verbatim/intended transcript pairs, we raise German disfluency F1 from 10\% to 79\% zero-shot from English-only training. Full English-only fine-tuning surpasses all baselines in verbatim accuracy, disfluency detection, and intended-mode quality across both languages. We introduce supervised cross-attention finetuning improving word-level timestamps on disfluent speech beyond forced alignment baselines. Finally we introduce a new task: verbatimize, enabling scalable creation/enrichment of speech corpora with high quality canonical verbatim transcriptions.
\end{abstract}

\section{Introduction}
\label{sec:intro}

Speech transcription requires a policy decision that most training pipelines leave implicit: should the output preserve what was said \emph{verbatim}, including filled pauses, repetitions, self-corrections, cut-offs, and paralinguistic sounds (e.g., ``I i [uh] li- like intended transcripts'') or should it produce \emph{intended} text that retains only the speaker's fluent content (``I like intended transcripts'')? Intended transcripts improve downstream text processing and readability~\cite{liu2024cos2w}, while verbatim transcripts preserve clinically and linguistically meaningful structure in spontaneous speech~\cite{shriberg1994preliminaries}. Disfluency patterns serve as diagnostic markers in neurological and speech disorders~\cite{clinical_disfluency2021,process_berns,zusag2023careful}, paralinguistic annotations improve naturalness in expressive speech synthesis~\cite{he2025emilia,disfluencyspeech2024,liao2024spontts,nvspeech2025}, and people who stutter experience systematic ASR failures including
truncations and elevated error rates~\cite{stutterzero2025,lea2023stutter_asr,sep28k_whisper2024}.

Despite the importance of this distinction, large-scale ASR systems are typically trained on data that mixes both conventions without a control signal. Whisper~\cite{radford2023robust}, trained on 680{,}000 hours of heterogeneously annotated data, may omit, inconsistently transcribe, or hallucinate disfluencies depending on acoustic context~\cite{koenecke2024careless}. Annotated disfluency corpora remain scarce, particularly beyond English~\cite{mujtaba2024inclusive,data_scarcity1,banno2025hesitation}, and large multilingual datasets~\cite{he2025emilia,granary} typically derive their labels from ASR pipelines trained on similarly heterogeneous data. Transcription style thus becomes an \emph{uncontrolled latent variable}: the same hesitation may be emitted, omitted, or rendered in non-canonical surface forms (um, uuumm, uhmm...) depending on surrounding context complicating downstream analysis.

This latent policy ambiguity causes a cascade of measurable failures. First, it produces \emph{decoding instability}: beam search yields substantially different transcripts for the same audio on disfluent speech, with median character-level beam divergence around 15\% (\autoref{sec:beam_divergence}), because the model inherently maintains competing transcription strategies with no signal for which to select. Second, it \emph{confounds evaluation}: we show that standard word error rate (WER) conflates content errors with style differences, and that up to 60\% of reported WER on conversational benchmarks reflects style mismatch rather than recognition failure (\autoref{sec:wer_confounding}). Third, it renders \emph{timing ill-defined}: if a model inconsistently transcribes word repetitions word-level timestamps become conceptually undefined rather than merely imprecise.

We argue that the core challenge is capability \emph{control}, not capability \emph{acquisition}. Prior work has shown that small-scale fine-tuning~\cite{crisperwhisper2024} and soft-prompt approaches~\cite{ma2023softprompt,gong2025prompting} can recover disfluencies from pretrained ASR models without architectural changes. We provide direct evidence: training only mode-tag embeddings with all encoder and decoder weights frozen raises German disfluency event F1 from 10\% to 79\%, a large capability gain with zero updates to the model's learned representations.

We activate this capability through three complementary mechanisms. \textbf{(1)~Explicit transcription policy via mode tags.} We introduce discrete decoder-prefix tokens that parameterize the transcription policy as a binary choice between verbatim and intended output. Trained with paired supervision-verbatim and intended transcripts for the same audio-and partitioned to reflect annotation coverage, mode tags disentangle \emph{what was said} from \emph{what to emit}. English-only verbatim supervision enables controling German output zero-shot, achieving 94\% disfluency event F1. \textbf{(2)~Supervised cross-attention for word-level timing.} Once the transcription policy is explicit and the output token sequence is stable, timing becomes well-defined. We directly supervise selected cross-attention heads with word-boundary targets, yielding 36\,ms mean absolute boundary error on read speech and 102\,ms on disfluent speech, for the first time outperforming forced-alignment baselines, which degrade to 142--200\,ms on disfluent input, with purely attention based methods. \textbf{(3)~Verbatimize: transcript-conditioned disfluency recovery.} Given audio and any intended transcript, verbatimize reconstructs a canonical verbatim version by inserting acoustically grounded disfluencies, raising rare-word recall from 6.8\% to 96.1\% and enabling cost efficient verbatim annotations at scale. \textbf{(4)~A language-agnostic evaluation framework.} We introduce an alignment-based protocol that automatically extracts typed disfluency labels from verbatim transcripts, disambiguates repetitions, and decomposes WER into content and style components, enabling consistent evaluation across languages and transcription conventions.

\section{Related Work}
\label{sec:related}

Table~\ref{tab:related} summarizes how existing approaches compare on transcription style control, timing, and multilingual support. We highlight the key gaps our work addresses.

\begin{table}[t]
\centering
\caption{Capability comparison with related work. \textbf{Verbatim (VB)}: produces verbatim transcripts with disfluencies. \textbf{Intended (IN)}: produces clean intended transcripts. \textbf{Controllable (CT)}: explicit switching between verbatim and intended modes. \textbf{Multilingual (ML)}: supports languages beyond English. \textbf{Sounds (SD)}: inline paralinguistic event tokens. \textbf{Timing (TG)}: word-level timestamps. {\normalfont$\sim$}\,=\,partial or unstable support.}
\label{tab:related}
\small
\setlength{\tabcolsep}{2.8pt}
\begin{tabular}{@{}l@{\hspace{6pt}}cccccc@{}}
\toprule
& {\textbf{VB}} 
& {\textbf{IN}} 
& {\textbf{CT}} 
& {\textbf{ML}} 
& {\textbf{SD}} 
& {\textbf{TG}} \\
\midrule
Whisper~\cite{radford2023robust}           & $\sim$ & \checkmark & --         & \checkmark & --         & $\sim$ \\
WhisperX~\cite{whisperx2023}               & $\sim$ & \checkmark & --         & \checkmark & --         & \checkmark\textsuperscript{a} \\
CrisperWhisper~\cite{crisperwhisper2024}   & \checkmark & --         & --         & $\sim$     & --         & \checkmark \\
Reverb~\cite{reverb2024}                   & \checkmark & \checkmark & \checkmark\textsuperscript{b} & -- & -- & -- \\
Soft-prompt~\cite{ma2023softprompt,gong2025prompting} & \checkmark & -- & -- & -- & -- & -- \\
Disfluency systems~\cite{mihajlik2024nonlexical,horii2022disfluency,futami2023streaming_joint}  & \checkmark & --         & --         & $\sim$     & $\sim$     & -- \\
Post-hoc detection~\cite{amann2024augmenting,zhu2022filler,kouzelis2023weakly}                  & $\sim$     & --         & --         & $\sim$     & --         & \checkmark \\
NVSpeech~\cite{nvspeech2025}               & --         & --         & --         & --         & \checkmark & -- \\
Disfluency removal~\cite{lou2020disfluency,stutterzero2025,liu2024cos2w}                        & --         & \checkmark & --         & $\sim$     & --         & -- \\
\midrule
\textbf{Ours}                              & \checkmark & \checkmark & \checkmark & \checkmark & \checkmark & \checkmark \\
\bottomrule
\end{tabular}
\vspace{2pt}
{\footnotesize
\textsuperscript{a}Via forced alignment
\textsuperscript{b}Continuous style parameter (English only)}
\end{table}

\textbf{Transcription style.} Most ASR systems produce transcripts under a single implicit policy, and approaches that target verbatim output typically do so unidirectionally-without bidirectional control between modes. Prompting and Soft-prompt approaches ~\cite{ma2023softprompt, gong2025prompting} can elicit disfluencies from foundation models but do not address stable switching under paired supervision. CrisperWhisper~\cite{crisperwhisper2024} fine-tunes Whisper on English verbatim data with tokenizer modifications that require multilingual verbatim training data for effective cross-lingual transfer, which is generally unavailable. Reverb~\cite{reverb2024} introduces a continuous verbatimicity parameter but is restricted to English and does not address timing. In contrast, our mode tags explicitly parameterize
the output policy and are trained on paired intended/verbatim
targets for the same audio, making style an explicit interface
rather than an uncontrolled latent variable

\textbf{Word-level timing.} Cross-attention heads in encoder-decoder models develop alignment-like behavior, and prior work extracts timestamps from these unsupervised patterns via dynamic time warping~\cite{radford2023robust,crisperwhisper2024}. The unreliable nature of these emergent alignments has motivated external components: WhisperX~\cite{whisperx2023} and Canary~\cite{canary2025} add separate forced aligners, improving accuracy at the cost of language-specific model dependencies. CrisperWhisper~\cite{crisperwhisper2024} improves attention-based alignment via explicit space tokens that function as word-boundary detectors, but this tokenizer modification limits cross-lingual generalization. A key observation motivates our approach: timing and transcription policy are coupled problems. When a model inconsistently transcribes disfluencies, cross-attention for those temporal regions has no stable target, making word-level timestamps conceptually ill-defined. We address both jointly: mode tags stabilize the output token sequence, and supervised cross-attention transforms emergent alignment into a trainable capability without external components or tokenizer changes.

\textbf{Disfluency modeling and evaluation.} End-to-end and multitask systems incorporate disfluency tags or non-lexical sound labels during training~\cite{mihajlik2024nonlexical,horii2022disfluency,futami2023streaming_joint,lian2023udm,mujtaba2024inclusive}, and inline speech sound event tokens have been explored in speech generation pipelines such as NVSpeech~\cite{nvspeech2025}, typically assuming a fixed labeling policy and targeting a narrower event inventory compared to our unified set of typed repetitions, cut-offs, filled pauses, and paralinguistic events. Post-hoc methods detect disfluencies after decoding: Zhu~et~al.~\cite{zhu2022filler} leverage ASR/VAD mismatch to propose timestamped filler candidates, Amann~et~al.~\cite{amann2024augmenting} localize open-set disfluency regions via modified CTC alignment and gap classification, and forced-alignment methods model transcript audio mismatch in disfluent speech~\cite{kouzelis2023weakly,zusag2023careful}. These provide useful primitives but depend on additional classification components and do not resolve the fundamental verbatim-versus-intended ambiguity. On the intended side, disfluency removal~\cite{lou2020disfluency,stutterzero2025,liu2024cos2w} is complementary to our work. Standard WER itself is part of the problem: professional transcripts of the same audio exhibit substantial stylistic variance~\cite{heuser2024quantification}, and reference choice alone can shift reported WER by several points~\cite{faria2022oracle,mcnamara2024style}. Our mode-conditioned approach complements these efforts: by explicitly controlling output style via tags, evaluation
can use style-matched references, isolating genuine errors from
style mismatch.

\section{The Transcription Policy Problem}
\label{sec:policy_problem}

Before presenting our method in Section~\ref{sec:method}, we establish that transcription policy ambiguity is not a theoretical concern but a measurable source of instability, evaluation error, and ill-defined timing in current ASR systems. 

\subsection{Policy ambiguity under mixed supervision}
\label{sec:dilution}

\begin{figure}[t]
\centering
\includegraphics[width=\columnwidth]{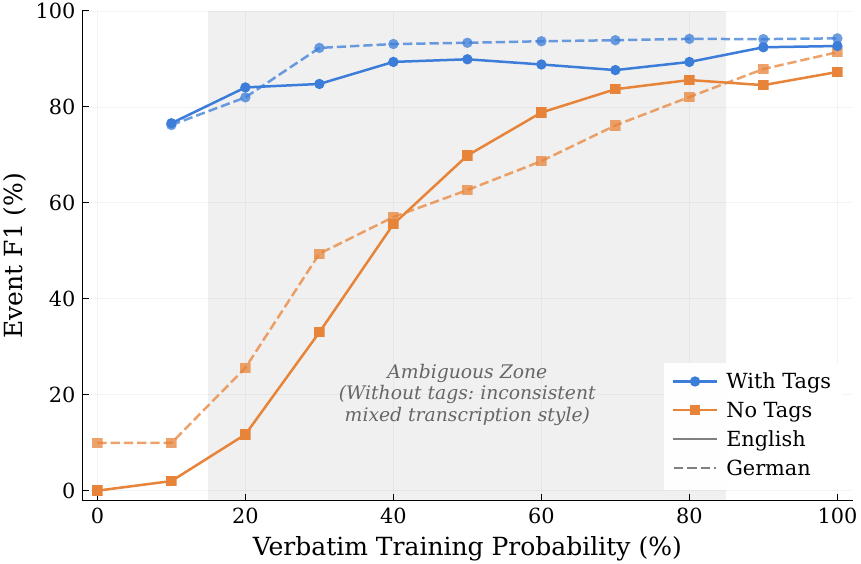}
\caption{Disfluency event F1 vs. verbatim training fraction: without mode tags, the ambiguous range (15–85\%) leads to unstable transcription behavior; with mode tags, performance remains stable across all data mixes. German shows the same pattern, even with no German verbatim training data.}
\label{fig:dilution}
\end{figure}

\begin{figure}[t]
\centering
\includegraphics[width=\columnwidth]{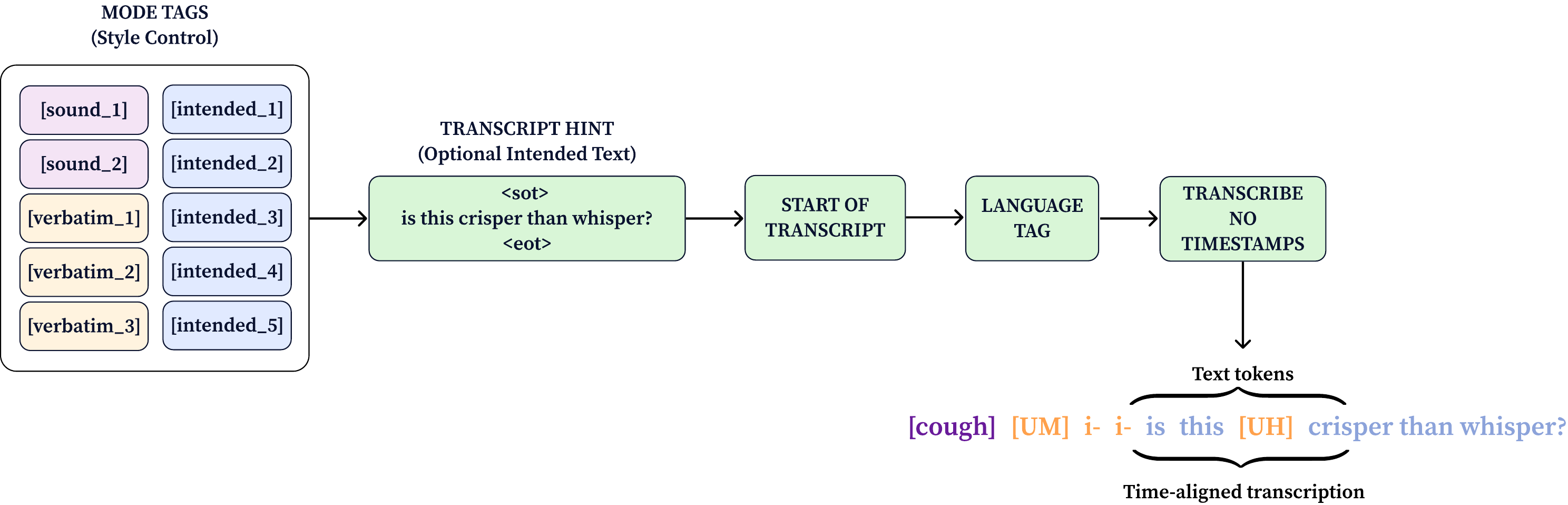}
\caption{Decoder prompt structure. Mode tags control tran-
scription style, followed by an optional intended transcript hint (for verbatimize), then standard Whisper tokens. The model outputs
time-aligned transcription in the prompted style }
\label{fig:decoder_prompt}
\end{figure}

To isolate the effect of annotation inconsistency, we vary the fraction of verbatim vs.\ intended annotated samples in the training data from 0\% to 100\%, training Whisper-medium models with and without mode tags (Figure~\ref{fig:decoder_prompt}) on each mixture for one epoch. Figure~\ref{fig:dilution} reports disfluency event F1 on English (DisfluencySpeech~\cite{disfluencyspeech2024}) and German (our German DisfluencySpeech evaluation set; Section~\ref{sec:data}).

Without mode tags, an ``ambiguous zone'' between 15--85\% fraction of verbatim data produces inconsistent behavior: disfluencies are sometimes transcribed and sometimes omitted, with no user-facing control. The model has learned competing transcription strategies from the mixed supervision and the output becomes unpredictable. With mode tags, event F1 remains around 80\% across all mixing ratios, including at 10\% verbatim where the untagged model achieves only 2\%. German exhibits the same pattern despite no German data in the training data.

\subsection{Decoding instability}
\label{sec:beam_divergence}

If transcription policy is unresolved, the ambiguity should manifest as decoding uncertainty-particularly on speech containing disfluencies, where verbatim and intended transcriptions diverge most. We measure this via \emph{beam divergence}: the median of the maximum pairwise character error rate (CER) among 10~beam hypotheses (temperature 0.7), averaged over samples.

Table~\ref{tab:beam_divergence} confirms this prediction. On disfluent speech samples from the AMI test set ~\cite{ami2005}, containing atleast one speech disfluency event (\autoref{sec:eval_protocol}), Whisper's beam divergence is 15.1\% CER, indicating that the model has internalized competing transcription strategies. NVIDIA Canary-1B~\cite{canary2025}, a 1B-parameter multilingual encoder-decoder model, shows comparable divergence (14.6\%). Our model with verbatim mode tags reduces divergence to 8.1\% (a 46\% reduction), and even intended-mode conditioning reduces it to 11.2\%, confirming that explicit policy directives resolve ambiguity in both directions. On clean speech (LibriSpeech), divergence is near zero for all Whisper-based models; Canary remains elevated at 5.0\%.

\input{tables/beam_divergence_table}

\subsection{Evaluation confounding}
\label{sec:wer_confounding}

Undefined transcription policy compromises not only model outputs but also their evaluation. When both the model prediction and the reference transcript carry implicit style choices, standard WER conflates content errors with style differences. To quantify this, we decompose WER into a \emph{content} component, measured by the content loss rate (CLR)-the fraction of intended-reference words absent from the prediction-and a \emph{style} component (WER\,$-$\,CLR), capturing differences due to disfluency inclusion ( fillers, repetitions, stutters, false starts).

Figure~\ref{fig:wer_decomp} presents this decomposition for TED-LIUM and AMI, two widely used benchmarks from the Open ASR Leaderboard~\cite{leaderboard}. Two findings emerge. First, verbatim and intended reference transcripts of the \emph{same audio}, which share 100\% of content words by construction (\autoref{sec:paired_construction}), disagree by 3.7\% WER on TED-LIUM and 12.1\% on AMI. This disagreement is entirely stylistic. Second, this style difference propagates into model evaluation: Whisper's reported WER on TED-LIUM shifts from 5.7\% to 7.3\% depending solely on reference choice, while its content loss is identical (a\,=\,3.8\%). On AMI, approximately 60\% of reported WER reflects style mismatch rather than information loss. Closer inspection reveals that TED-LIUM transcribes repetitions but omits fillers and cut-offs while AMI transcribes fully verbatim without sounds. Further incosistencies in the popular Leaderboard have recently been noted in ~\cite{artificialanalysis2026aawer} suggesting that style-aware evaluation should become standard practice for conversational ASR benchmarks ~\cite{mcnamara2024style}. Mode-conditioned transcription offers a path forward: by making output style an explicit parameter, evaluation can use style-matched references, verbatim predictions against verbatim ground truth, intended against intended isolating genuine recognition errors from annotation convention differences, further CLR is a stable metric to evaluate content indenpendent of style.

\input{figures/wer_decomp_figure}

\subsection{Timing is ill-defined under policy ambiguity}
\label{sec:timing_illdefined}

The three failures above have a direct consequence for word-level timing. If the model may or may not emit a disfluency, the cross-attention distribution for that temporal region has no consistent target. Base Whisper's cross-attention yields 203\,ms mean absolute boundary error (MAE) on read speech (TIMIT) and 568\,ms on disfluent speech (FluencyBank)-effectively unusable. Reliable timing therefore requires a stable output token sequence and explicit alignment supervision; our method addresses both.

\section{Method}
\label{sec:method}

We present three mechanisms to unambiguously activate latent capabilities in large-scale ASR model: explicit style control via mode tags, supervised cross-attention for word-level timing, and transcript-conditioned disfluency recovery. All experiments initialize from the official OpenAI Whisper-medium checkpoint~\cite{radford2023robust}.
We extend Whisper's vocabulary with atomic tokens for phenomena absent from the original tokenizer. 

\textbf{Filled pauses} (2~tokens): \texttt{[uh]}, \texttt{[um]}. 

\textbf{Paralinguistic sound events} (12~tokens): \texttt{[laughter]}, \texttt{[cough]}, \texttt{[sigh]}, \texttt{[breath]}, \texttt{[lipsmack]}, \texttt{[sniff]}, \texttt{[throat\text{-}clearing]}, \texttt{[yawn]}, \texttt{[noise]}, \texttt{[crying]}, \texttt{[scream]}, \texttt{[sneeze]}. 

\textbf{Mode tags} (10~tokens): \texttt{[verbatim\_1,2,3]}, \texttt{[sound\_1,2]},\texttt{[intended\_1,2,3,4,5]}. 

\textbf{Verbatimize delimiters} (2~tokens): \texttt{<sot>} and \texttt{<eot>}. Following prior work~\cite{nvspeech2025,crisperwhisper2024}, sound and filler tokens appear at their temporal position within the transcript.

\subsection{Activating style control: mode tags}
\label{sec:mode_tags}

\subsubsection{Mode tag design}

We condition the decoder on an explicit transcription policy $s \in \{\text{verbatim}, \text{intended}\}$ via prefix tokens prepended to the decoder input. In \emph{verbatim mode}, the model transcribes all audible content including disfluencies and paralinguistic sounds. In \emph{intended mode}, it produces clean fluent text, omitting reparanda, interregna, and paralinguistic events~\cite{shriberg1994preliminaries,levelt1983}:

\begin{center}
\small
\begin{tabular}{@{}r@{\hspace{4pt}}p{0.75\columnwidth}@{}}
\textbf{Verbatim:} & I i \texttt{[uh]} \texttt{[laughter]} li- like, yeah \texttt{[um]} i like intended transcripts. \\
\textbf{Intended:} & I like intended transcripts. \\
\end{tabular}
\end{center}

\noindent We use five tokens per mode \texttt{[verbatim\_1,2,3]} and  \texttt{[sound\_1,2]} for verbatim, \texttt{[intended\_1..5]}) for intended; multiple prefix tokens provide a stronger conditioning signal than a single token and enable compositional control, as described below.

\subsubsection{Coverage-aware tag partitioning}
\label{sec:tag_partition}

Real-world training data is heterogeneously annotated: some corpora mark disfluencies but not sounds, others include injected sound events but lack reliable disfluency labels. Naively training on all data with a single set of verbatim tags produces contradictory gradients, the model is penalized for predicting a sound event in one sample and for omitting the identical event in another, solely because annotations are incomplete rather than because the events are absent.

We resolve this by partitioning the verbatim tags by annotation coverage. Tags \texttt{[verbatim\_1..3]} control disfluency transcription (fillers, repetitions, cut-offs) and tags \texttt{[sound\_1,2]} control paralinguistic sound detection. During training, each sample receives only the tags matching its annotation coverage: verbatim corpora with disfluency labels but no sound annotations use \texttt{[verbatim\_1..3]}; clean samples with injected sound events (Section~\ref{sec:paired_construction}) use only \texttt{[sound\_1,2]}; fully annotated corpora use all five. This partitioning gives the model a learnable signal that distinguishes incomplete label coverage from absent phenomena, allowing it to attribute missing annotations to the data rather than to the audio.

\subsection{Vocabulary extension}
\label{sec:vocab}
\subsubsection{Paired training}

For each training sample, we randomly select verbatim or intended mode, prepend the corresponding tags, and supervise with the matching transcript. The same audio thus appears under both transcription policies across training, distinguished only by the mode tags. This teaches a clean separation of roles: the encoder representations determine all available transcription content (words, disfluencies, sounds), while the mode tags determine the \emph{emission policy}-which content to include in the output.

\subsection{Activating timing: supervised cross-attention}
\label{sec:timing_method}

With transcription policy resolved and token outputs stable, word-level timing becomes well-defined. Yet cross-attention alignment remains emergent and unreliable in the base model (Section~\ref{sec:timing_illdefined}). By directly training selected cross-attention alignment heads in a supervised fashion, we enable systematic improvements in timing accuracy.

\subsubsection{Training objective}

Let $\mathcal{W} = \{(w_i, s_i, e_i)\}_{i=1}^{N}$ denote a transcript of $N$~words, where word~$w_i$ spans the interval $[s_i, e_i]$ in seconds. The encoder produces $F$~frames at 20\,ms resolution ($F = 1500$ for 30\,s input). The decoder generates $T$~tokens $(y_1, \ldots, y_T)$, each belonging to exactly one word~$w_{\pi(t)}$ via the mapping function $\pi(t)$. For each token~$y_t$ and attention head~$h$, cross-attention yields a distribution $\mathbf{a}_t^{(h)} \in \mathbb{R}^F$ over encoder frames.

For token~$y_t$ belonging to word~$w_{\pi(t)}$ with time span $[s_{\pi(t)}, e_{\pi(t)}]$, we construct a binary target $\mathbf{g}_t \in \{0,1\}^{F}$ that is one for frames within the word span and zero elsewhere:
\begin{equation}
g_t(f) = \mathbf{1}\bigl[\lfloor s_{\pi(t)}/\Delta \rfloor \leq f \leq \lfloor e_{\pi(t)}/\Delta \rfloor\bigr],
\label{eq:gt}
\end{equation}
where $\Delta = 0.02$\,s is the frame shift. In other words, $\mathbf{g}_t$ is one inside the span of word~$w_{\pi(t)}$ and zero everywhere else. We select the $k{=}10$ cross-attention heads whose unsupervised attention patterns best correlate with ground-truth alignments on TIMIT~\cite{timit1993}, adding heads greedily by marginal improvement. Let $\mathcal{H}$ denote this head subset. During training, we compute their averaged attention:
\begin{equation}
\bar{\mathbf{a}}_t = \frac{1}{|\mathcal{H}|} \sum_{h \in \mathcal{H}} \mathbf{a}_t^{(h)},
\label{eq:avg_attn}
\end{equation}
and minimize the mean cosine distance between averaged attention and ground truth:
\begin{equation}
\mathcal{L}_{\mathrm{timing}} = \frac{1}{T}\sum_{t=1}^{T} \left( 1 - \frac{\bar{\mathbf{a}}_t \cdot \mathbf{g}_t}{\lVert \bar{\mathbf{a}}_t \rVert \, \lVert \mathbf{g}_t \rVert} \right).
\label{eq:timing_loss}
\end{equation}
Cosine distance is scale-invariant, comparing distribution shape rather than magnitude. Optimizing the \emph{averaged} attention is critical: it allows individual heads to specialize in complementary acoustic aspects of the predicted token, whereas per-head loss optimizes each head in isolation, preventing cooperative specialization. We validate this empirically in Section~\ref{sec:results_timing}.

\subsubsection{Word-level timing extraction at inference}
\label{sec:inference_timing}

Whisper's tokenizer merges preceding whitespace into word-initial tokens (e.g., ``\textvisiblespace meeting''), causing untrained cross-attention to conflate preceding pauses with word onsets. Prior work addresses this via explicit space tokens~\cite{crisperwhisper2024}, requiring tokenizer modifications that limit cross-lingual generalization and slow inference. We instead decouple pause detection from the attention mechanism at inference time, requiring no architectural changes.
\newline
\noindent\textbf{Energy-based pause model.} We construct a virtual blank (pause) emission probability from the utterance-normalized frame-wise mel energy:
\begin{equation}
p(\text{blank} \mid f) = \max\bigl(0,\; \min\bigl(1,\; 1 - \tfrac{E_{\text{mel}}(f) - p_{10}}{p_{90} - p_{10}}\bigr)\bigr),
\label{eq:blank}
\end{equation}
where $p_{10}$ and $p_{90}$ are the 10th and 90th energy percentiles of the current utterance.
\newline
\noindent\textbf{Attention sharpening.} We sharpen attention distributions via temperature scaling: $\mathrm{attn}'_{ij} = (\mathrm{attn}_{ij})^\tau \,/\, \sum_k (\mathrm{attn}_{ik})^\tau$, with $\tau{=}3$ at inference.

\noindent\textbf{Word-level aggregation.} Sharpened token-level attention is aggregated into word-level emission log-probabilities via log-sum-exp over the tokens comprising each word.

\noindent\textbf{Viterbi alignment.} We align words to frames using Viterbi decoding over a state space that strictly alternates between virtual blank states and word states. Blank states absorb pauses when acoustic energy is low; transitions into word states follow the aggregated attention. 

\subsection{Activating transcript-conditioned generation: verbatimize}
\label{sec:verbatimize}

Explicit policy control enables a further capability: retroactive annotation. High-quality intended transcripts are abundant, but verbatim transcripts with consistent disfluency annotation are scarce and costly to create. \emph{Verbatimize} bridges this gap: given audio and any intended transcript, the model reconstructs a canonical verbatim version by copying the intended content while inserting acoustically grounded disfluencies at their temporal locations.

\noindent\textbf{Prompt format.} The decoder input contains the intended transcript wrapped in delimiters, preceded by verbatim mode tags:
\[
\texttt{[verbatim\_1..5]}\ \texttt{<sot>}\ \text{intended\_text}\ \texttt{<eot>}
\]
The models task is to reconstruct the verbatim transcript as output, inserting disfluency markers (\texttt{[uh]}, \texttt{[um]}, repetitions, cut-offs, sound tags) at audio-detected positions.

\noindent\textbf{Training modes.} In \emph{standard verbatimize}, the prompt contains the intended transcript and the target is the verbatim transcript. We validate that the intended transcript forms a subsequence of the verbatim transcript for valid training pairs, ensuring that disfluencies are strictly insertions. In \emph{verbatimize-copy}, clean samples use identical prompt and target, teaching the model not to hallucinate disfluencies when none are present in the audio.

\noindent\textbf{Casing perturbation.} To encourage the model to attend closely to the prompt rather than relying solely on acoustics, we apply symmetric casing perturbation: 1--3 content words (selected by length, excluding stop words) are randomly uppercased in \emph{both} prompt and target. This forces the model to consult the prompt for exact spelling, while relying on the audio for disfluency localization.

\subsection{Staged training}
\label{sec:staged_training}

To preserve Whisper's pretrained multilingual capabilities, we train in two stages. \textbf{Stage~1} freezes all pretrained parameters and trains only the newly added token embeddings (mode tags, delimiters, sound markers) for 1~epoch. This stage also serves as the latent-capability experiment reported in Section~\ref{sec:results_style}: even without any updates to the encoder or decoder, mode-tag embeddings alone substantially activate verbatim transcription. \textbf{Stage~2} unfreezes the decoder and continues training with a reduced learning rate. The pre-learned tag embeddings provide stable initialization and mitigate catastrophic forgetting.

\section{Data}
\label{sec:data}

Our training procedure requires paired transcripts: each audio sample needs both a verbatim and an intended reference.

\subsection{Corpora}
\label{sec:corpora}

Table~\ref{tab:corpora} summarizes all datasets and their roles. For English verbatim data, we use the ICSI Meeting Corpus~\cite{icsi2004} (72\,h), the AMI Meeting Corpus~\cite{ami2005} (100\,h), CORAAL~\cite{coraal2021} (150\,h), and the National Speech Corpus~\cite{nsc_corpus} (2{,}000\,h), however we only train on 40h subset with high quality verbatim labels. For clean read speech, we use LibriSpeech~\cite{librispeech2015} (960\,h) and Common Voice~\cite{commonvoice2020} (multilingual). We additionally contribute a \textbf{German DisfluencySpeech (GDS)} evaluation set\footnote{Models, Data and evaluation code: \url{https://github.com/nyrahealth/CrisperWhisper}}: following the English DisfluencySpeech design~\cite{disfluencyspeech2024}, native German speakers recorded 202~utterances containing controlled disfluencies (fillers, repetitions, cut-offs, vocal sounds, self-corrections), each with parallel verbatim and intended transcripts. For sound augmentation, we use VocalSound~\cite{vocalsound2022} (21k~samples) and Nonspeech7k~\cite{nonspeech7k2023} (7k~samples).

\begin{table}[t]
\centering
\caption{Datasets used in this work. \textbf{Role}: Train = training, Eval = evaluation. \textbf{Annotation}: V = verbatim transcripts, C = clean transcripts, T = word-level timestamps.}
\label{tab:corpora}
\small
\setlength{\tabcolsep}{3pt}
\begin{tabular}{@{}llrcl@{}}
\toprule
\textbf{Corpus} & \textbf{Lang.} & \textbf{Hours} & \textbf{Ann.} & \textbf{Role} \\
\midrule
ICSI~\cite{icsi2004}                    & EN    & 72    & V   & Train, Eval \\
AMI~\cite{ami2005}                      & EN    & 100   & V   & Train \\
CORAAL~\cite{coraal2021}                & EN    & 150   & V   & Train \\
NSC~\cite{nsc_corpus}                   & EN    & 2{,}000 & V & Train \\
LibriSpeech~\cite{librispeech2015}      & EN    & 960   & C   & Train \\
Common Voice~\cite{commonvoice2020}     & Multi & var.  & C   & Train \\
\midrule
DisfluencySpeech~\cite{disfluencyspeech2024} & EN & 10  & V, T & Eval \\
FluencyBank~\cite{fluencybank2024}      & EN    & 4     & V, T & Eval \\
TIMIT~\cite{timit1993}                  & EN    & 5     & C, T & Eval \\
Thorsten~\cite{thorsten_dataset}        & DE    & 23    & C, T & Eval \\
GDS (ours)          & DE    & 2     & V   & Eval \\
\bottomrule
\end{tabular}
\end{table}

\subsection{Paired transcript construction and preprocessing}
\label{sec:paired_construction}

For corpora with verbatim annotations (ICSI, AMI, CORAAL, NSC), we generate intended transcripts using GPT-4o~\cite{gpt-4o} by stripping fillers and sound tags, collapsing repetitions and false starts to their repairs, removing fragments, and reformatting numbers and dates for readability. For clean corpora, verbatim and intended transcripts are treated as identical since these recordings hardly contain any disfluencies. To provide supervision for paralinguistic sound detection, we augment 30{,}000~clean samples by injecting 1--2~vocal events from VocalSound or Nonspeech7k into pauses exceeding 200\,ms, identified via Montreal Forced Aligner (MFA)~\cite{mfa2017} alignments. Sound tags are added to the verbatim transcript only; augmented samples receive only sound-detection tags (\texttt{[sound\_1,2]}), consistent with coverage-aware partitioning (Section~\ref{sec:tag_partition}).

For quality filtering on clean corpora, we transcribe each sample with Whisper-medium and retain only those achieving ${<}3\%$ WER against the Whisper prediction, removing mislabeled or acoustically degraded samples. Word-level timestamps for cross-attention supervision are generated via MFA~\cite{mfa2017,rousso2024alignment} for 500 hours. All transcripts are normalized to a canonical form: filled pauses as \texttt{[uh]}/\texttt{[um]}, fragments/cutoffs with trailing hyphens, sound events as single-token tags at their temporal position, and numbers in spoken-word form.

To evaluate verbatimize (Section~\ref{sec:verbatimize}), we use GPT-4o to identify ICSI transcripts containing rare or domain-specific words (technical jargon, proper nouns) unlikely to be recoverable from acoustics alone. We manually verify 1{,}843 rare-word occurrences across 1{,}342 samples (1{,}591 unique word types), confirming that none appear elsewhere in the training corpus.

\section{Experimental Setup}
\label{sec:experimental_setup}

\subsection{Evaluation datasets}

We evaluate three capabilities on separate benchmarks. \textbf{Disfluency detection:} DisfluencySpeech~\cite{disfluencyspeech2024} (English, $n{=}4{,}707$) and German DisfluencySpeech (German, $n{=}202$). \textbf{Word-level timing:} TIMIT~\cite{timit1993} (English, read speech), FluencyBank~\cite{fluencybank2024} (English, disfluent speech), and Thorsten~\cite{thorsten_dataset} (German, read speech). \textbf{Verbatimize:} the ICSI rare-word evaluation set ($n{=}1{,}342$; Section~\ref{sec:paired_construction}).

\subsection{Disfluency evaluation protocol}
\label{sec:eval_protocol}

Our canonical verbatim format enables automatic extraction of gold disfluency labels via edit-distance alignment between predicted and reference verbatim transcripts. Matched tokens are classified as fluent; unmatched tokens are assigned a disfluency type deterministically by surface pattern: a trailing hyphen indicates a \textsc{cutoff}, a bracketed filler indicates a \textsc{filler}, any other bracketed token indicates a \textsc{sound}, a consecutive duplicate preceding a matched token indicates a \textsc{repetition}, and any remaining unmatched sequence is classified as \textsc{other} (e.g.\ self-corrections, restarts, or novel disfluencies not captured by the above rules).

\noindent\textbf{Repetition disambiguation.} Consecutive identical tokens (e.g., ``the the the problem'') create alignment ambiguity, since repeated and fluent instances are equally valid edit-distance targets. Following Lou and Johnson~\cite{lou2020disfluency}, we use modified alignment costs that prefer matching fluent words over disfluent copies. The last instance before a non-duplicate is tagged as fluent; all preceding duplicates are tagged as repetitions. Table~\ref{tab:rep_example} illustrates this procedure.

\begin{table}[t]
\centering
\caption{Repetition disambiguation via modified edit-distance alignment. The fluent instances of ``I'' and ``the'' are anchored first (last duplicate before a non-duplicate). The prediction incorrectly produces a cut-off ``I-'' where the reference has a repetition ``I'' (false positive for \textsc{cutoff}, false negative for \textsc{repetition}) and deletes one repetition of ``the'' (false negative for \textsc{repetition}).}
\label{tab:rep_example}
\small
\setlength{\tabcolsep}{3.5pt}
\begin{tabular}{@{}lccccccc@{}}
\toprule
\textbf{Reference} & I & I & {[um]} & the & the & problem & is \\
\textbf{Label}      & \textsc{rep} & \textsc{flu} & \textsc{fill} & \textsc{rep} & \textsc{flu} & \textsc{flu} & \textsc{flu} \\
\midrule
\textbf{Prediction} & I- & I & {[um]} & - & the & problem & is \\
\textbf{Label}      & \textsc{cut} & \textsc{flu} & \textsc{fill} & - & \textsc{flu} & \textsc{flu} & \textsc{flu} \\
\midrule
\textbf{Result}     & \makecell{\textsc{fp}\textsubscript{cut}\\\textsc{fn}\textsubscript{rep}} & \textsc{tp}\textsubscript{flu} & \textsc{tp}\textsubscript{fill} & \textsc{fn}\textsubscript{rep} & \textsc{tp}\textsubscript{flu} & \textsc{tp}\textsubscript{flu} & \textsc{tp}\textsubscript{flu} \\
\bottomrule
\end{tabular}
\end{table}

\subsection{Metrics}
\label{sec:metrics}

All metrics are micro-averaged at corpus level. Text is lowercased and punctuation removed before computation.

\noindent\textbf{Transcription quality.} We report verbatim word error rate (vWER) and character error rate (vCER) against verbatim references, and intended word error rate (iWER) against clean references, decomposed into substitution rate (iSR, indicating misrecognition), deletion rate (iDR, indicating over-removal of content), and insertion rate (iIR, indicating retained disfluencies or hallucinations).

\noindent\textbf{Disfluency detection.} Per-type F1 scores for fillers (fF1), sounds (sF1), cut-offs (cF1), and repetitions (rF1), computed via the alignment procedure above. Cut-off F1 requires a trailing hyphen at the aligned position; lenient cut-off recall (lcR) accepts any token at gold cut-off positions. Event F1 (eF1) is the type-agnostic F1 over all disfluency types.

\noindent\textbf{Verbatimize.} Content loss rate (CLR): the fraction of intended-reference words not recovered in the prediction. Rare-word recall (RWR): the fraction of manually verified rare words correctly preserved.

\noindent\textbf{Timing.} Mean absolute boundary error (MAE): the mean of start and end offset errors across all aligned word pairs, in milliseconds. F1@$X$\,ms: the fraction of aligned pairs where both start and end offsets fall within a tolerance of $X$\,ms (collars: 50, 100, 200\,ms), following~\cite{crisperwhisper2024}.

\subsection{Baselines}
\label{sec:baselines}

We compare against Whisper-medium~\cite{radford2023robust}, Canary-1B~\cite{canary2025}, Reverb~\cite{reverb2024} (English only), AssemblyAI Universal-3-Pro~\cite{assemblyai}, and CrisperWhisper~\cite{crisperwhisper2024}. For timing, we additionally compare against MFA~\cite{mfa2017} and WhisperX~\cite{whisperx2023}.

\subsection{Training details}
\label{sec:training_details}

We fine-tune Whisper-medium using Hugging Face Transformers. Stage~1: new embeddings only, 1~epoch, lr $5{\times}10^{-4}$. Stage~2: decoder unfrozen, 3~epochs, lr $1{\times}10^{-5}$. Batch size 160 with 4-step gradient accumulation, fp16, single NVIDIA A100. The timing loss $\mathcal{L}_{\text{timing}}$ (Equation~\ref{eq:timing_loss}) is weighted at 0.2 relative to cross-entropy.

\section{Results}
\label{sec:results}

We present results organized by the three capabilities activated in Section~\ref{sec:method}: style control, word-level timing and verbatimize.

\subsection{Style control: latent capability and cross-lingual transfer}
\label{sec:results_style}

Tables~\ref{tab:disfluency} and~\ref{tab:intended} present disfluency detection and intended transcript quality on English and German. We report our model at two key stages: \textbf{S1} (Stage~1: frozen model, only tag embeddings trained) and \textbf{FT\textsubscript{0}} (Stage~2: full decoder fine-tuning, zero German verbatim training data). For FT\textsubscript{0}, we report results both with and without mode tags to isolate the effect of explicit policy control. Adding 10k~German verbatim samples (from a proprietary inhouse dataset) yields marginal gains over FT\textsubscript{0} (e.g., 93.8\%$\to$95.2\% German eF1); we report the fully trained model FT\textsubscript{all} as the final system.

\begin{table}[t]
\caption{Disfluency detection on English ($n{=}4{,}957$) and German ($n{=}202$). Best per column and language in bold. Reverb is English-only.}
\label{tab:disfluency}
\centering
\resizebox{\columnwidth}{!}{%
\begin{tabular}{@{}lc|ccccccc@{}}
\toprule
\textbf{Model} & \textbf{L} & \textbf{vWER}$\downarrow$ & \textbf{eF1}$\uparrow$ & \textbf{fF1}$\uparrow$ & \textbf{sF1}$\uparrow$ & \textbf{cF1}$\uparrow$ & \textbf{rF1}$\uparrow$ & \textbf{lcR}$\uparrow$ \\
\midrule
Whisper & EN & 10.9 & 12.0 & 14.3 & 0.0 & 4.8 & 7.6 & 29.6 \\
        & DE & 24.7 & 10.3 & 13.7 & 0.0 & 0.0 & 4.1 & 0.3 \\
\midrule
Canary-1B & EN & 9.6 & 19.9 & 24.2 & 0.0 & 0.0 & 30.4 & 33.7 \\
          & DE & 24.5 & 10.5 & 13.9 & 0.0 & 0.0 & 2.8 & 0.8 \\
\midrule
Reverb & EN & 5.2 & 86.3 & \textbf{93.3} & 54.2 & 33.5 & 86.5 & 83.3 \\
\midrule
AssemblyAI & EN & 4.9 & 85.8 & 92.5 & 68.8 & 17.0 & 84.8 & 71.0 \\
           & DE & 18.1 & 44.5 & 22.3 & 70.9 & 28.2 & 74.7 & 24.3 \\
\midrule
CrisperWhisper & EN & 6.4 & 73.2 & 81.7 & 0.0 & 0.0 & 68.0 & 73.2 \\
               & DE & 15.1 & 60.0 & 71.7 & 0.0 & 0.0 & 60.2 & 45.8 \\
\midrule
S1, with tags & EN & 9.1 & 53.0 & 60.6 & 26.8 & 17.6 & 45.4 & 88.3 \\
& DE & 12.9 & 78.9 & 78.6 & 51.5 & 22.8 & 82.7 & 45.6 \\
\midrule
FT\textsubscript{0}, with tags & EN & \textbf{4.0} & \textbf{90.7} & 90.1 & \textbf{87.3} & \textbf{43.2} & \textbf{87.2} & 95.4 \\
                          & DE & 5.1 & 93.8 & 95.3 & \textbf{89.5} & 81.5 & \textbf{88.5} & 89.9 \\
FT\textsubscript{0}, without tags & EN & 5.5 & 80.7 & 85.4 & 55.7 & 20.4 & 78.6 & 79.7 \\
                              & DE & 16.2 & 60.1 & 64.2 & 44.7 & 21.6 & 59.0 & 29.4 \\
\midrule
FT\textsubscript{10k}, with tags & EN & 4.8 & 87.9 & 88.2 & 88.2 & 39.1 & 87.1 & 95.1 \\
                            & DE & 4.5 & \textbf{95.0} & \textbf{97.0} & 89.7 & \textbf{88.7} & 86.0 & \textbf{96.2} \\
FT\textsubscript{10k}, without tags & EN & 4.8 & 83.9 & 80.0 & 63.2 & 17.2 & 71.3 & 78.0 \\
                                & DE & 13.1 & 74.2 & 78.3 & 62.0 & 66.9 & 65.4 & 59.0 \\
\bottomrule
\end{tabular}%
}
\end{table}

\begin{table}[t]
\caption{Intended transcript quality (same models and conditions as Table~\ref{tab:disfluency}). iIR dominates errors for untagged models, reflecting disfluency leakage into intended output.}
\label{tab:intended}
\centering{%
\begin{tabular}{@{}lc|cccc@{}}
\toprule
\textbf{Model} & \textbf{L} & \textbf{iWER}$\downarrow$ & \textbf{iSR}$\downarrow$ & \textbf{iDR}$\downarrow$ & \textbf{iIR}$\downarrow$ \\
\midrule
Whisper & EN & 14.2 & 2.6 & 1.4 & 10.1 \\
        & DE & 5.7 & 1.1 & 0.4 & 4.2 \\
\midrule
Canary-1B & EN & 15.4 & 2.6 & 1.1 & 11.6 \\
          & DE & \textbf{4.6} & \textbf{0.9} & 0.3 & 3.4 \\
\midrule
Reverb & EN & 11.5 & 2.9 & 2.2 & 6.4 \\
\midrule
AssemblyAI & EN & 21.1 & 3.3 & \textbf{0.7} & 17.1 \\
           & DE & 15.4 & 1.0 & 0.1 & 14.3 \\
\midrule
CrisperWhisper & EN & 20.5 & 2.9 & 1.0 & 16.5 \\
               & DE & 19.6 & 1.2 & \textbf{0.0} & 18.4 \\
\midrule
S1, with tags & EN & 12.3 & 2.4 & 2.0 & 7.9 \\
 & DE & 5.5 & 1.1 & 0.3 & 4.2 \\
\midrule
FT\textsubscript{0}, with tags & EN & 9.4 & 2.5 & 2.7 & 4.2 \\
                          & DE & 4.9 & 1.0 & 0.2 & 3.6 \\
FT\textsubscript{0}, without tags & EN & 18.8 & \textbf{2.1} & 0.8 & 15.9 \\
                              & DE & 17.0 & 1.0 & 0.1 & 15.9 \\
\midrule
FT\textsubscript{10k}, with tags & EN & \textbf{9.3} & 2.5 & 2.8 & \textbf{4.1} \\
                            & DE & 4.7 & 1.0 & 0.3 & \textbf{3.4} \\
FT\textsubscript{10k}, without tags & EN & 19.2 & 2.2 & 0.8 & 16.3 \\
                               & DE & 23.2 & 1.0 & 0.1 & 22.1 \\
\bottomrule
\end{tabular}%
}
\end{table}

\noindent\textbf{Evidence for latent capability.} Unmodified Whisper detects disfluencies at 12.0\% event F1 on English and 10.3\% on German, despite no explicit verbatim training. Stage~1 (S1), which freezes the entire model and trains \emph{only} mode-tag embeddings, raises event F1 to 53.0\% on English and 78.9\% on German-a large capability gain with zero weight updates to encoder or decoder. Full decoder fine-tuning with zero German verbatim data (FT\textsubscript{0}) further improves performance to 90.7\% English and 93.8\% German. The progression 12\%$\to$53\%$\to$91\% (English) and 10\%$\to$79\%$\to$94\% (German) demonstrates that the capability is progressively \emph{unlocked} rather than learned from scratch. 

\noindent\textbf{Mode tags are the critical structural signal.} The tags-versus-no-tags comparison on FT\textsubscript{0} shows the benefit of explicit policy control. With tags, the model achieves 93.8\% German event F1 zero-shot; without tags, the same model reaches only 60.1\%. This 34 percentage point gap, which could be arbitarily extreme by adjusting the data mixture ~\ref{fig:dilution} demonstrates that target-language data cannot substitute for explicit policy parameterization: further mode tags enable the model to reliably leverage its multilingual pretraining for cross-lingual transfer. 

\noindent\textbf{Intended quality is preserved.} The iWER decomposition (Table~\ref{tab:intended}) reveals that insertion rate (iIR) dominates intended-mode errors. Models without tags leak disfluencies into intended output (iIR <15\%), whereas mode tags strongly suppress this leakage (iIR 3--4\%). Substitution and deletion rates remain stable below 3\% and 1\% respectively, confirming that explicit style control does not compromise recognition accuracy.

\subsection{Word-level timing}
\label{sec:results_timing}

Table~\ref{tab:timing} compares timing accuracy across three evaluation sets: TIMIT (English, read speech), FluencyBank (English, disfluent speech), and Thorsten (German, read speech). We report results both without and with inference-time sharpening ($\tau{=}3$).

\begin{table*}[h]
\caption{Word-level timing comparison. MAE (ms; lower is better) and F1 (\%; higher is better) at boundary tolerance thresholds. "+s" indicates inference-time sharpening ($\tau{=}3$). "ah": averaged-head loss (Eq.~\ref{eq:timing_loss}); "ph": per-head loss. Best per column in bold.}
\label{tab:timing}
\centering
\small
\setlength{\tabcolsep}{3pt}
\begin{tabular}{@{}l|cccc|cccc|cccc@{}}
\toprule
& \multicolumn{4}{c|}{\textbf{TIMIT (EN, read)}} & \multicolumn{4}{c|}{\textbf{FluencyBank (EN, disfluent)}} & \multicolumn{4}{c}{\textbf{Thorsten (DE, read)}} \\
\textbf{Method} & \textbf{MAE}$\downarrow$ & \textbf{F1@50} & \textbf{F1@100} & \textbf{F1@200} & \textbf{MAE}$\downarrow$ & \textbf{F1@50} & \textbf{F1@100} & \textbf{F1@200} & \textbf{MAE}$\downarrow$ & \textbf{F1@50} & \textbf{F1@100} & \textbf{F1@200} \\
\midrule
Base Whisper           & 203 & 0.4 & 4.0 & 32.9 & 568 & 0.6 & 5.5 & 26.3 & 218 & 0.5 & 2.8 & 25.7 \\
Base Whisper+s         & 197 & 1.5 & 7.3 & 33.5 & 944 & 0.9 & 6.3 & 25.9 & 222 & 0.4 & 2.8 & 25.3 \\
\midrule
MFA                    & \textbf{19} & \textbf{84.6} & \textbf{97.0} & \textbf{99.8} & 142 & \textbf{56.1} & \textbf{75.5} & \textbf{85.1} & --- & --- & --- & --- \\
WhisperX               & 66 & 12.5 & 60.7 & 97.8 & 200 & 8.9 & 48.0 & 80.2 & 89 & 9.8 & 51.2 & 91.5 \\
CrisperWhisper         & 59 & 41.5 & 70.1 & 92.8 & 162 & 23.1 & 48.5 & 70.8 & 67 & 34.2 & 63.9 & 86.3 \\
CrisperWhisper+s       & 47 & 50.1 & 81.8 & 95.9 & 122 & 29.9 & 54.8 & 75.6 & 57 & 38.8 & 72.1 & 90.2 \\
Canary-1B              & 78 & 14.9 & 46.1 & 94.0 & 166 & 9.5 & 31.7 & 75.8 & 103 & 5.6 & 29.9 & 84.1 \\
\midrule
Ours (ph)              & 57 & 42.8 & 71.2 & 93.2 & 159 & 24.0 & 49.2 & 71.5 & 65 & 35.1 & 64.8 & 87.0 \\
Ours (ah)              & 49 & 50.5 & 76.0 & 95.2 & 141 & 26.7 & 52.0 & 72.9 & 60 & 38.6 & 68.0 & 88.3 \\
Ours (ph)+s            & 45 & 51.2 & 82.9 & 96.4 & 119 & 30.8 & 55.7 & 76.3 & 56 & 39.6 & 73.0 & 90.9 \\
Ours (ah)+s            & \textbf{36} & \textbf{64.0} & \textbf{87.4} & \textbf{97.3} & \textbf{102} & \textbf{38.6} & \textbf{61.1} & \textbf{80.3} & \textbf{55} & \textbf{42.8} & \textbf{74.0} & \textbf{90.9} \\
\bottomrule
\end{tabular}
\end{table*}
\noindent\textbf{Supervision transforms attention into a precise aligner.} Without timing supervision, base Whisper's cross-attention is too diffuse for boundary extraction: 203\,ms MAE on TIMIT, 568\,ms on FluencyBank. Applying sharpening without supervision helps marginally on read speech (203$\to$197\,ms) but \emph{worsens} performance on disfluent speech (568$\to$944\,ms), because concentrating mass on incorrectly positioned attention amplifies errors. Supervised training reduces MAE by 5--6$\times$ (to 36--102\,ms), demonstrating that attention based alignment methods can outperform strong force alignment baselines.

\noindent\textbf{Supervised attention outperforms forced alignment on disfluent speech.} MFA achieves the lowest MAE on read speech (19\,ms on TIMIT) but degrades to 142\,ms on FluencyBank---worse than both CrisperWhisper+s (122\,ms) and our method (102\,ms). Forced alignment relies on canonical pronunciation models that fail on repetitions, repairs, and filled pauses. WhisperX shows the same limitation: 66\,ms on TIMIT but 200\,ms on FluencyBank. By training cross-attention heads alongside transcription policy allows our method to extracts boundary information from non-canonical speech naturally.

\noindent\textbf{Cross-lingual timing transfer.} On Thorsten (German), our English-trained model achieves 55\,ms MAE with zero German timing supervision, outperforming WhisperX (89\,ms) and Canary-1B (103\,ms). CrisperWhisper+s reaches 57\,ms on this dataset but was extensively trained on german. Our attention-based method generalizes well across languages without language-specific components, while WhisperX's wav2vec\,2.0 German model proves less accurate than its English counterpart.

\noindent\textbf{Averaged-head loss enables cooperative specialization.} Computing the timing loss on the averaged attention (ah) consistently outperforms per-head loss (ph): 36 vs.\ 45\,ms on TIMIT, 102 vs.\ 119\,ms on FluencyBank. Averaging before the loss allows individual heads to attend to complementary aspects of each token; per-head loss prevents this cooperation. We saw that MAE decreases consistently as more heads are added (78$\to$39\,ms on TIMIT with 1$\to$10 heads), with diminishing returns beyond~8.

\noindent\textbf{Inference-time sharpening.} Raising cross-attention to the power $\tau{=}3$ consistently improves all attention-based methods. The learned distributions are correctly centered but slightly diffuse around boundaries; sharpening suppresses tail mass without shifting centers. The technique requires no retraining and appears broadly applicable.

\subsection{Verbatimize}
\label{sec:results_verbatimize}

Table~\ref{tab:verbatimize} evaluates verbatimize on the ICSI rare-word evaluation set ($n{=}1{,}342$). Without verbatimize, the model transcribes from acoustics alone (9.4\% CLR, 6.8\% RWR). Verbatimize training reduces CLR to 1.5\% and raises RWR to 94.1\%, demonstrating that the model learns to faithfully copy prompted content while inserting disfluencies from the audio. Casing perturbation further improves all metrics (CLR 1.3\%, RWR 96.1\%) by forcing closer attention to exact surface forms. 
\begin{table}[h]
\centering
\caption{Verbatimize results on the ICSI rare-word set ($n{=}1{,}342$). B and B+V share weights; B uses standard verbatim mode, B+V adds the intended transcript as prompt. B+V+C adds casing perturbation during training.}
\label{tab:verbatimize}
\small
\begin{tabular}{@{}lcccc@{}}
\toprule
\textbf{Model} & \textbf{CLR}$\downarrow$ & \textbf{RWR}$\uparrow$ & \textbf{vWER}$\downarrow$ & \textbf{vCER}$\downarrow$ \\
\midrule
B       & 9.4 & 6.8  & 12.8 & 7.7 \\
B+V     & 1.5 & 94.1 & 4.6  & 4.3 \\
B+V+C   & \textbf{1.3} & \textbf{96.1} & \textbf{4.4} & \textbf{4.2} \\
\bottomrule
\end{tabular}
\end{table}
The near equality of vWER and vCER for trained models indicates predominantly single-character residual errors; manual inspection confirms these are largely cut-off transcription ambiguities (e.g., ``\texttt{t- trr- transformer}'' vs.\ ``\texttt{d- tr- transformer}''), inherently ambiguous.

\section{Discussion and Conclusion}
\label{sec:discussion}

Across Sections~\ref{sec:results_style}--\ref{sec:results_verbatimize}, we find a consistent pattern: pretrained ASR models often already contain the needed capability, but require an explicit interface to reliably express it. The frozen-model experiment (S1) is the clearest example: training only 27 new token embeddings, with no updates to the 764M encoder/decoder parameters, raises German event F1 from 10\% to 79\%. Likewise, when a large-scale pretrained model behaves inconsistently, the right intervention may be a control signal rather than more data: our dilution experiment (Figure~\ref{fig:dilution}) shows that adding verbatim data without tags does not resolve the ambiguous zone, whereas tags resolve it even with only 10\% verbatim data.

We also address the data scarcity that constrains disfluency-aware speech processing. Annotated disfluency corpora remain scarce~\cite{mujtaba2024inclusive,liao2024spontts,data_scarcity1,banno2025hesitation}, and large multilingual datasets~\cite{he2025emilia,granary} often derive labels from ASR pipelines trained on heterogeneously annotated data, propagating labeling errors across generations of corpora and models. Verbatimize helps break this cycle by decoupling verbatim transcription from intended-transcript accuracy: the 96.1\% rare-word recall indicates that the model preserves prompted content while inserting acoustically grounded disfluency events, enabling scalable corpus enrichment without costly manual re-annotation or hybrid pipelines where one component produces high quality intended transcripts that verbatimize upgrades to highly accurate canonical verbatim transcripts.

Overall, transcription style is an uncontrolled latent variable in current large-scale ASR, contributing to decoding instability, evaluation confounding, and ill-defined timing. Discrete mode tags make transcription policy controllable; supervised cross-attention provides precise word-level timing once outputs are stable; and verbatimize enables scalable data enrichment. The jump from 10\% to 94\% zero-shot German disfluency F1 and to 79\% with zero weight updates supports the view that mode-specific transcription largely requires activation rather than acquisition, as quantified by our language-agnostic evaluation framework.

\

\noindent\textbf{Limitations.} We evaluated cross-lingual transfer exclusively on German, a typologically close language to English; evaluation on distant language pairs is needed. The German evaluation set was recorded by the authors, who are not trained speakers, so disfluencies may be less naturalistic than in the wild. Timing supervision relies on MFA-generated alignments rather than hand-annotated boundaries for most training data.

\noindent\textbf{Future work.} Tag-based conditioning could be unified with speaker diarization, using speaker embeddings as additional decoder prefix tokens to jointly control style and attribution. For timing, training dedicated alignment heads from scratch rather than repurposing pretrained cross-attention and more sophisticated losses with VAD or pause detection could improve precision. Finally, Verbatimize with beam search and heuristic filtering enables large-scale bootstrapping of consistent verbatim corpora, unlocking high quality weakly supervised data for better verbatim ASR, speech analysis and expressive TTS.

\section{Generative AI Use Disclosure}
Some code used in the experiments was written with help from a
coding assistant (Claude by Anthropic). The experiments were run manually and results were manually verified. Generative AI
was also used in the formatting of tables and plots.
The paper was manually written. The authors assume full responsibility and
accountability for the content of this submission.

\bibliographystyle{IEEEtran}
\bibliography{verbatim_paper}

\end{document}

%% file: tables/beam_divergence_table.tex
\begin{table}[t]
\centering
\caption{Beam divergence (median of maximum pairwise CER across 10 beam hypotheses, temperature 0.7). High values indicate that the model produces substantially different transcripts for the same audio across decoding paths.}
\label{tab:beam_divergence}
\small
\begin{tabular}{@{}lcc@{}}
\toprule
& \multicolumn{2}{c}{\textbf{Divergence (\%)~$\downarrow$}} \\
\cmidrule(l){2-3}
\textbf{Model} & \textbf{AMI (disfluent)} & \textbf{LibriSpeech (clean)} \\
\midrule
Canary-1B~\cite{canary2025}       & 14.6          & 5.0 \\
Whisper~\cite{radford2023robust}   & 15.1          & 0.0 \\
Ours (intended)               & 11.2          & 1.0 \\
Ours (verbatim)               & \textbf{8.1}  & 0.0 \\
\bottomrule
\end{tabular}
\end{table}

%% file: figures/wer_decomp_figure.tex
\begin{figure}[t]
\centering
\begin{tikzpicture}
\begin{axis}[
    ybar stacked,
    bar width=16pt,
    width=4.6cm,
    height=5.2cm,
    ylabel={Word Error Rate (\%)},
    title={\textbf{TED-LIUM}},
    symbolic x coords={VI, WV, WI},
    xtick=data,
    xticklabels={V{\footnotesize$\to$}I, W{\footnotesize$\to$}V, W{\footnotesize$\to$}I},
    x tick label style={font=\small},
    ymin=0, ymax=22,
    enlarge x limits=0.35,
    ymajorgrids=true,
    grid style={dashed, gray!40},
    legend to name=shared_wer_legend,
    legend columns=-1,
    legend style={font=\footnotesize, draw=none, /tikz/every even column/.append style={column sep=0.5cm}},
    legend cell align={left},
]
\addplot+[ybar, fill=blue!55, draw=blue!75,
          nodes near coords,
          point meta=explicit,
          every node near coord/.append style={font=\scriptsize, text=white, anchor=center, yshift=-0.5ex}]
    coordinates {(VI, 0.00) [] (WV, 3.77) [3.8] (WI, 3.77) [3.8]};

\addplot+[ybar, fill=red!45, draw=red!65,
          nodes near coords={\pgfmathprintnumber\pgfplotspointmeta},
          point meta=explicit,
          every node near coord/.append style={font=\scriptsize, text=white, anchor=center, yshift=-0.5ex}]
    coordinates {(VI, 3.69) [3.7] (WV, 2.88) [1.9] (WI, 3.53) [3.5]};

\addplot+[ybar, fill=none, draw=none,
          nodes near coords,
          point meta=explicit,
          every node near coord/.append style={font=\footnotesize\bfseries, text=black, anchor=south, yshift=1pt}]
    coordinates {(VI, 0) [3.7] (WV, 0) [5.7] (WI, 0) [7.3]};

\legend{Content (CLR), Style (WER$-$CLR)}
\end{axis}
\end{tikzpicture}%
\hfill%
\begin{tikzpicture}
\begin{axis}[
    ybar stacked,
    bar width=16pt,
    width=4.6cm,
    height=5.2cm,
    title={\textbf{AMI}},
    symbolic x coords={VI, WV, WI},
    xtick=data,
    xticklabels={V{\footnotesize$\to$}I, W{\footnotesize$\to$}V, W{\footnotesize$\to$}I},
    x tick label style={font=\small},
    ymin=0, ymax=22,
    enlarge x limits=0.35,
    ymajorgrids=true,
    grid style={dashed, gray!40},
    yticklabels={,,},
]
\addplot+[ybar, fill=blue!55, draw=blue!75,
          nodes near coords,
          point meta=explicit,
          every node near coord/.append style={font=\scriptsize, text=white, anchor=center, yshift=-0.5ex}]
    coordinates {(VI, 0.00) [] (WV, 7.69) [7.7] (WI, 7.69) [7.7]};

\addplot+[ybar, fill=red!45, draw=red!65,
          nodes near coords={\pgfmathprintnumber\pgfplotspointmeta},
          point meta=explicit,
          every node near coord/.append style={font=\scriptsize, text=white, anchor=center, yshift=-0.5ex}]
    coordinates {(VI, 12.11) [12.1] (WV, 11.56) [11.6] (WI, 11.04) [11.0]};

\addplot+[ybar, fill=none, draw=none,
          nodes near coords,
          point meta=explicit,
          every node near coord/.append style={font=\footnotesize\bfseries, text=black, anchor=south, yshift=1pt}]
    coordinates {(VI, 0) [12.1] (WV, 0) [19.3] (WI, 0) [18.7]};

\end{axis}
\end{tikzpicture}

\vspace{-2pt}
\ref{shared_wer_legend}
\caption{WER decomposition into content loss (CLR, blue) and style mismatch (red) on TED-LIUM and AMI. V\,$\to$\,I = verbatim vs.\ intended reference; W\,$\to$\,V = Whisper vs.\ verbatim; W\,$\to$\,I = Whisper vs.\ intended. Bold totals show reported WER.}
\label{fig:wer_decomp}
\end{figure}

%% file: verbatim_paper.bib
@inproceedings{mfa2017,
  title={Montreal Forced Aligner: Trainable Text-Speech Alignment Using Kaldi},
  author={McAuliffe, Michael and Socolof, Michaela and Mihuc, Sarah and Wagner, Michael and Sonderegger, Morgan},
  booktitle={Proceedings of Interspeech},
  pages={498--502},
  year={2017}
}

@article{mcnamara2024style,
  title={Style-agnostic evaluation of ASR using multiple reference transcripts},
  author={McNamara, Quinten and del R{\'\i}o Fern{\'a}ndez, Miguel {\'A}ngel and Bhandari, Nishchal and Ratajczak, Martin and Chen, Danny and Miller, Corey and Jett{\'e}, Mig{\"u}el},
  journal={arXiv preprint arXiv:2412.07937},
  year={2024}
}

@inproceedings{lou2020disfluency,
  title={End-to-End Speech Recognition and Disfluency Removal},
  author={Lou, Paria Jamshid and Johnson, Mark},
  booktitle={Findings of the Association for Computational Linguistics: EMNLP 2020},
  pages={2051--2061},
  year={2020}
}

@article{amann2024augmenting,
  title={Augmenting Automatic Speech Recognition Models with Disfluency Detection},
  author={Amann, Robin and Li, Zhaolin and Bruno, Barbara and Niehues, Jan},
  journal={arXiv preprint arXiv:2409.10177},
  year={2024}
}

@inproceedings{heuser2024quantification,
author = {Heuser, Annika and Kendall, Tyler and Rio, Miguel and McNamara, Quinn and Bhandari, Nishchal and Miller, Corey and Jetté, Migüel},
year = {2024},
month = {09},
pages = {4538-4542},
title = {Quantification of stylistic differences in human- and ASR-produced transcripts of African American English},
doi = {10.21437/Interspeech.2024-2300}
}

@inproceedings{faria2022oracle,
  title={Toward Zero Oracle Word Error Rate on the Switchboard Benchmark},
  author={Faria, Arlo and Janin, Adam and Riedhammer, Korbinian and Adkoli, Sidhi},
  booktitle={Proc. INTERSPEECH},
  pages={3973--3977},
  year={2022}
}

@misc{assemblyai,
  author       = {{AssemblyAI}},
  title        = {Introducing {Universal-3 Pro}: A New Class of Speech Language Model Optimized for Voice {AI}},
  year         = {2026},
  howpublished = {\url{https://www.assemblyai.com/blog/introducing-universal-3-pro}},
}

@inproceedings{librispeech2015,
  title={Librispeech: An ASR corpus based on public domain audio books},
  author={Panayotov, Vassil and Chen, Guoguo and Povey, Daniel and Khudanpur, Sanjeev},
  booktitle={Proceedings of ICASSP},
  pages={5206--5210},
  year={2015}
}

@article{radford2023robust,
  title={Robust speech recognition via large-scale weak supervision},
  author={Radford, Alec and Kim, Jong Wook and Xu, Tao and Brockman, Greg and McLeavey, Christine and Sutskever, Ilya},
  journal={International Conference on Machine Learning},
  pages={28492--28518},
  year={2023},
  publisher={PMLR}
}

@inproceedings{crisperwhisper2024,
  title={{CrisperWhisper}: Accurate Timestamps on Verbatim Speech Transcriptions},
  author={Wagner, Laurin and Thallinger, Bernhard and Zusag, Mario},
  booktitle={Proceedings of INTERSPEECH},
  year={2024}
}

@article{nvspeech2025,
  title={{NVSpeech}: An Integrated and Scalable Pipeline for Human-Like Speech Modeling with Paralinguistic Vocalizations},
  author={Liao, Huan and Ni, Qinke and Wang, Yuancheng and Lu, Yiheng and Zhan, Haoyue and Xie, Pengyuan and Zhang, Qiang and Wu, Zhizheng},
  journal={arXiv preprint arXiv:2508.04195},
  year={2025}
}

@inproceedings{whisperx2023,
  title={{WhisperX}: Time-accurate speech transcription of long-form audio},
  author={Bain, Max and Huh, Jaesung and Han, Tengda and Zisserman, Andrew},
  booktitle={Proceedings of INTERSPEECH},
  year={2023}
}

@inproceedings{rousso2024alignment,
author = {Rousso, Rotem and Cohen, Eyal and Keshet, Joseph and Chodroff, Eleanor},
year = {2024},
month = {09},
pages = {1525-1529},
title = {Tradition or Innovation: A Comparison of Modern ASR Methods for Forced Alignment},
doi = {10.21437/Interspeech.2024-429}
}

@inproceedings{vocalsound2022,
  title={{VocalSound}: A dataset for improving human vocal sounds recognition},
  author={Gong, Yuan and Yu, Jianbo and Glass, James},
  booktitle={IEEE International Conference on Acoustics, Speech and Signal Processing (ICASSP)},
  pages={151--155},
  year={2022}
}

@article{nonspeech7k2023,
  title={{Nonspeech7k} dataset: Classification and analysis of human non-speech sound},
  author={Rashid, M. M. and Li, G. and Du, C.},
  journal={IET Signal Processing},
  volume={17},
  number={6},
  pages={e12233},
  year={2023}
}

@article{disfluencyspeech2024,
  title={{DisfluencySpeech}: Single-Speaker Conversational Speech Dataset with Paralanguage},
  author={Wang, Kyra and Herremans, Dorien},
  journal={arXiv preprint arXiv:2406.08820},
  year={2024},
  note={10h studio-quality single-speaker dataset derived from Switchboard with filled pauses, discourse markers, restarts, and non-speech sounds; provides parallel verbatim/intended transcripts at three levels of disfluency removal}
}

@article{shriberg1994preliminaries,
  title={Preliminaries to a theory of speech disfluencies},
  author={Shriberg, Elizabeth Ellen},
  journal={PhD dissertation, University of California, Berkeley},
  year={1994}
}

@article{fluencybank2024,
  title={{FluencyBank} Timestamped: An Updated Data Set for Disfluency Detection and Automatic Intended Speech Recognition},
  author={Romana, Amrit and Niu, Minxue and Perez, Matthew and Mower Provost, Emily},
  journal={Journal of Speech, Language, and Hearing Research},
  volume={67},
  number={11},
  pages={4203--4215},
  year={2024},
  doi={10.1044/2024_JSLHR-24-00070},
  note={Provides word-level timestamps and disfluency labels for stuttered speech; enables benchmarking of disfluency detection models}
}

@article{timit1993,
  title={The {DARPA TIMIT} acoustic-phonetic continuous speech corpus ({TIMIT})},
  author={Garofolo, John S and Lamel, Lori F and Fisher, William M and Fiscus, Jonathan G and Pallett, David S},
  journal={Linguistic Data Consortium},
  year={1993}
}

@inproceedings{icsi2004,
  title={The {ICSI} meeting corpus},
  author={Janin, Adam and Baron, Don and Edwards, Jane and Ellis, Dan and Gelbart, David and Morgan, Nelson and Peskin, Barbara and Pfau, Thilo and Shriberg, Elizabeth and Stolcke, Andreas and others},
  booktitle={IEEE International Conference on Acoustics, Speech, and Signal Processing},
  year={2003}
}

@inproceedings{ami2005,
  title={The {AMI} meeting corpus},
  author={Carletta, Jean and Ashby, Simone and Bourban, Sebastien and Flynn, Mike and Guillemot, Mael and Hain, Thomas and Kadlec, Jaroslav and Karaiskos, Vasilis and Kraaij, Wessel and Kronenthal, Melissa and others},
  booktitle={International Conference on Methods and Techniques in Behavioral Research},
  year={2005}
}

@misc{coraal2021,
  author    = {Kendall, Tyler and Farrington, Charlie},
  title     = {The Corpus of Regional African American Language},
  year      = {2023},
  version   = {2023.06},
  address   = {Eugene, OR},
  publisher = {The Online Resources for African American Language Project},
  doi       = {10.7264/1ad5-6t35},
  url       = {https://doi.org/10.7264/1ad5-6t35}
}

@inproceedings{commonvoice2020,
  title={{Common Voice}: A massively-multilingual speech corpus},
  author={Ardila, Rosana and Branson, Megan and Davis, Kelly and Henretty, Michael and Kohler, Michael and Meyer, Josh and Morais, Reuben and Saunders, Lindsay and Tyers, Francis M and Weber, Gregor},
  booktitle={Proceedings of the 12th Language Resources and Evaluation Conference},
  pages={4211--4215},
  year={2020}
}

@article{clinical_disfluency2021,
  title={Clinical applications of disfluency analysis in neurological populations},
  author={Ash, Sharon and Grossman, Murray},
  journal={Aphasiology},
  year={2021}
}

@inproceedings{zusag2023careful,
  title={Careful Whisper -- Leveraging Advances in Automatic Speech Recognition for Robust and Interpretable Aphasia Subtype Classification},
  author={Zusag, Mario and Wagner, Laurin and Bloder, Theresa},
  booktitle={Proceedings of INTERSPEECH},
  pages={3933--3937},
  year={2023},
  note={Achieves human-level aphasia classification by combining CTC (acoustic/verbatim) and encoder-decoder (clean) ASR outputs}
}

@misc{reverb2024,
      title={Reverb: Open-Source ASR and Diarization from Rev}, 
      author={Nishchal Bhandari and Danny Chen and Miguel Ángel del Río Fernández and Natalie Delworth and Jennifer Drexler Fox and Migüel Jetté and Quinten McNamara and Corey Miller and Ondřej Novotný and Ján Profant and Nan Qin and Martin Ratajczak and Jean-Philippe Robichaud},
      year={2025},
      eprint={2410.03930},
      archivePrefix={arXiv},
      primaryClass={cs.CL},
      url={https://arxiv.org/abs/2410.03930}, 
}

@inproceedings{ma2023softprompt,
  title={Adapting an {ASR} Foundation Model for Spoken Language Assessment},
  author={Ma, Rao and Qian, Mengjie and Gales, Mark J. F. and Knill, Kate M.},
  booktitle={Proceedings of INTERSPEECH},
  year={2023},
  note={Soft prompt tuning for verbatim output; English-only, unidirectional, no timing}
}

@inproceedings{horii2022disfluency,
  title={End-to-End Spontaneous Speech Recognition Using Disfluency Labeling},
  author={Horii, Koharu and Fukuda, Meiko and Ohta, Kengo and Nishimura, Ryota and Ogawa, Atsunori and Kitaoka, Norihide},
  booktitle={Proceedings of INTERSPEECH},
  year={2022},
  note={Joint ASR with disfluency tags for spontaneous speech}
}

@inproceedings{futami2023streaming_joint,
  title={Streaming Joint Speech Recognition and Disfluency Detection},
  author={Futami, Hayato and Tsunoo, Emiru and Shibata, Kentaro and Kashiwagi, Yosuke and Okuda, Takao and Arora, Siddhant and Watanabe, Shinji},
  booktitle={Proceedings of ICASSP},
  year={2023},
  note={Streaming encoder-decoder joint ASR and disfluency detection (multitask / tag-based variants)}
}

@inproceedings{lian2023udm,
  title={Unconstrained Dysfluency Modeling for Dysfluent Speech Transcription and Detection},
  author={Lian, Jiachen and Feng, Carly and Farooqi, Naasir and Li, Steve and Kashyap, Anshul and Cho, Cheol Jun and Wu, Peter and Netzorg, Robbie and Li, Tingle and Anumanchipalli, Gopala Krishna},
  booktitle={Proceedings of the IEEE Automatic Speech Recognition and Understanding Workshop (ASRU)},
  year={2023},
  note={Defines and models dysfluent speech transcription and detection; introduces simulated dysfluent dataset (VCTK++)},
  eprint={2312.12810},
  archivePrefix={arXiv},
  primaryClass={cs.CL}
}

@inproceedings{kouzelis2023weakly,
  title={Weakly-Supervised Forced Alignment of Disfluent Speech Using Phoneme-Level Modeling},
  author={Kouzelis, Theodoros and Paraskevopoulos, Georgios and Katsamanis, Athanasios and Katsouros, Vassilis},
  booktitle={Proceedings of INTERSPEECH},
  year={2023},
  eprint={2306.00996},
  archivePrefix={arXiv},
  primaryClass={cs.CL},
  note={Forced alignment under transcript--audio mismatch due to disfluencies; avoids requiring fully verbatim transcripts}
}

@article{lea2023stutter_asr,
  title={From User Perceptions to Technical Improvement: Enabling People Who Stutter to Better Use Speech Recognition},
  author={Lea, Colin and Huang, Zifang and Tooley, Lauren and Narain, Jayashree and Yee, Dianna and Georgiou, Panayiotis and Tran, Tien Dung and Bigham, Jeffrey P. and Findlater, Leah},
  journal={ACM Transactions on Accessible Computing},
  volume={15},
  number={2},
  pages={1--27},
  year={2023},
  note={Reports 19.8\% WER and 23.8\% truncation rate on production ASR for people who stutter}
}

@inproceedings{sep28k_whisper2024,
author = {Sridhar, Charan and Wu, Shaomei},
year = {2025},
month = {08},
pages = {3753-3757},
title = {J-j-j-just Stutter: Benchmarking Whisper's Performance Disparities on Different Stuttering Patterns},
doi = {10.21437/Interspeech.2025-2700}
}

@article{stutterzero2025,
  title={{StutterZero} and {StutterFormer}: End-to-End Speech Conversion for Stuttering Transcription and Correction},
  author={Xu, Qianheng},
  journal={IEEE Access},
  volume={13},
  pages={208773--208787},
  year={2025},
  note={Converts stuttered waveforms to fluent waveforms; 24--28\% WER reduction vs Whisper-Medium on converted output}
}

@inproceedings{koenecke2024careless,
  title={Careless Whisper: Speech-to-Text Hallucination Harms},
  author={Koenecke, Allison and Choi, Anna S. G. and Mei, Katrina X. and Schellmann, Hilke and Sloane, Meredith},
  booktitle={Proceedings of ACM FAccT},
  year={2024}
}

@article{gong2025prompting,
  title={Prompting {Whisper} for Improved Verbatim Transcription and End-to-end Miscue Detection},
  author={Smith, Griffin Dietz and Yee, Dianna and Chen, Jennifer King and Findlater, Leah},
  journal={arXiv preprint arXiv:2505.23627},
  year={2025}
}

@article{liao2024spontts,
  title={Spontaneous Style Text-to-Speech Synthesis with Controllable Spontaneous Behaviors Based on Language Models},
  author={Li, Weiqin and Yang, Peiji and Zhong, Yicheng and Zhou, Yixuan and Wang, Zhisheng and Wu, Zhiyong and Wu, Xixin and Meng, Helen},
  journal={arXiv preprint arXiv:2407.13509},
  year={2024},
  note={Addresses data scarcity for spontaneous speech TTS}
}

@inproceedings{liu2024cos2w,
  title={Recording for Eyes, Not Echoing to Ears: Contextualized Spoken-to-Written Conversion of {ASR} Transcripts},
  author={Liu, Jiaqing and Deng, Chong and Zhang, Qinglin and Zhou, Shilin and Chen, Qian and Yu, Hai and Wang, Wen},
  booktitle={Proceedings of AAAI},
  year={2025},
  note={Demonstrates that downstream NLP tasks benefit from converting verbatim ASR to clean text}
}

@article{he2025emilia,
  title={Emilia: A Large-Scale, Extensive, Multilingual, and Diverse Dataset for Speech Generation},
  author={He, Haorui and Shang, Zengqiang and Wang, Chaoren and Li, Xuyuan and Gu, Yicheng and Hua, Hua and Liu, Liwei and Yang, Chen and Li, Jiaqi and Shi, Peiyang and Wang, Yuancheng and Chen, Kai and Zhang, Pengyuan and Wu, Zhizheng},
  journal={IEEE Transactions on Audio, Speech and Language Processing},
  year={2025},
  note={101k+ hours of spontaneous speech; models produce more human-like output than audiobook-trained models}
}

@INPROCEEDINGS{process_berns,
  author={Thallinger, Bernhard and Wagner, Laurin and Bloder, Theresa and Zusag, Mario},
  booktitle={ICASSP 2025 - 2025 IEEE International Conference on Acoustics, Speech and Signal Processing (ICASSP)}, 
  title={A Multi-Stage Feature Pipeline on Timestamped Speech Transcriptions for Dementia Assessment}, 
  year={2025},
  volume={},
  number={},
  pages={1-2},
  doi={10.1109/ICASSP49660.2025.10889023}}

@inproceedings{mujtaba2024inclusive,
  title={Inclusive {ASR} for Disfluent Speech: Cascaded Large-Scale Self-Supervised Learning with Targeted Fine-Tuning and Data Augmentation},
  author={Mujtaba, Dena and Mahapatra, Nihar R. and Arney, Megan and Yaruss, J. Scott and Herring, Caryn and Bin, Jia},
  booktitle={Proceedings of INTERSPEECH},
  year={2024},
  note={States ``a critical barrier to progress is the scarcity of large, annotated disfluent speech datasets''}
}

@inproceedings{mihajlik2024nonlexical,
  title={On Disfluency and Non-lexical Sound Labeling for End-to-end Automatic Speech Recognition},
  author={Mihajlik, Peter and Meng, Yan and K{\'a}d{\'a}r, M{\'a}t{\'e} and Linke, Julian and Schuppler, Barbara and Mady, Katalin},
  booktitle={Proceedings of INTERSPEECH},
  year={2024},
  note={Systematic comparison of disfluency/non-lexical labeling strategies on conversational Hungarian and Austrian German; distinguishes filled pauses from meaningful backchannels}
}

@inproceedings{banno2025hesitation,
  title={Acoustically Precise Hesitation Tagging Is Essential for End-to-End Verbatim Transcription Systems},
  author={Lin, Jhen-Ke and Lu, Hao-Chien and Wang, Chung-Chun and Lin, Hong-Yun and Chen, Berlin},
  booktitle={arXiv preprint arXiv:2506.04076},
  year={2025},
  note={States manual annotation is prohibitively expensive; used LLM labeling at \$5 for entire dataset vs high cost of human annotators}
}

@inproceedings{zhu2022filler,
  title={Filler Word Detection and Classification: A Dataset and Benchmark},
  author={Zhu, Ge and Caceres, Juan-Pablo and Salamon, Justin},
  booktitle={Proceedings of INTERSPEECH},
  year={2022},
  note={Introduces PodcastFillers (145h, 35K fillers) and an ASR+VAD+classifier pipeline for timestamped filler detection in podcasts}
}

@inproceedings{data_scarcity1,
author = {Mohapatra, Payal and Pandey, Akash and Islam, Bashima and Zhu, Qi},
title = {Speech Disfluency Detection with Contextual Representation and Data Distillation},
year = {2022},
isbn = {9781450394031},
publisher = {Association for Computing Machinery},
address = {New York, NY, USA},
url = {https://doi.org/10.1145/3539490.3539601},
doi = {10.1145/3539490.3539601},
booktitle = {Proceedings of the 1st ACM International Workshop on Intelligent Acoustic Systems and Applications},
pages = {19–24},
numpages = {6},
location = {Portland, OR, USA},
series = {IASA '22}
}

@article{canary2025,
  title={{Canary-1B-v2} \& {Parakeet-TDT-0.6B-v3}: Efficient and High-Performance Models for Multilingual {ASR} and {AST}},
  author={Sekoyan, Monica and Koluguri, Nithin Rao and Tadevosyan, Nune and Zelasko, Piotr and Bartley, Travis and Karpov, Nikolay and Balam, Jagadeesh and Ginsburg, Boris},
  journal={arXiv preprint arXiv:2509.14128},
  year={2025},
  note={NVIDIA's 1B multilingual encoder-decoder ASR trained on 1.7M hours; uses NFA for timestamps due to cross-attention limitations}
}

@misc{granary,
      title={Granary: Speech Recognition and Translation Dataset in 25 European Languages}, 
      author={Nithin Rao Koluguri and Monica Sekoyan and George Zelenfroynd and Sasha Meister and Shuoyang Ding and Sofia Kostandian and He Huang and Nikolay Karpov and Jagadeesh Balam and Vitaly Lavrukhin and Yifan Peng and Sara Papi and Marco Gaido and Alessio Brutti and Boris Ginsburg},
      year={2025},
      eprint={2505.13404},
      archivePrefix={arXiv},
      primaryClass={cs.CL},
      url={https://arxiv.org/abs/2505.13404}, 
}

@inproceedings{nsc_corpus,
  title     = {Building the Singapore English National Speech Corpus},
  author    = {Jia Xin Koh and Aqilah Mislan and Kevin Khoo and Brian Ang and Wilson Ang and Charmaine Ng and Ying-Ying Tan},
  year      = {2019},
  booktitle = {Interspeech 2019},
  pages     = {321--325},
  doi       = {10.21437/Interspeech.2019-1525},
  issn      = {2958-1796},
}

@misc{gpt-4o,
  title={Hello {GPT-4o}},
  author={OpenAI},
  year={2024},
  howpublished={\url{https://openai.com/index/hello-gpt-4o/}}
}

@article{levelt1983,
  title={Monitoring and self-repair in speech},
  author={Levelt, Willem J. M.},
  journal={Cognition},
  volume={14},
  number={1},
  pages={41--104},
  year={1983},
  publisher={Elsevier}
}

@misc{leaderboard,
  title={Open {ASR} Leaderboard},
  author={{Hugging Face}},
  year={2024},
  howpublished={\url{https://huggingface.co/spaces/hf-audio/open_asr_leaderboard}}
}

@dataset{thorsten_dataset,
  author={M{\"u}ller, Thorsten and Kreutz, Dominik},
  title={{Thorsten} - Open German Voice (Neutral) Dataset},
  year={2021},
  publisher={Zenodo},
  doi={10.5281/zenodo.5525342}
}

@article{artificialanalysis2026aawer,
  title     = {{AA-WER} v2.0: Speech to Text Accuracy Benchmark},
  author    = {{Artificial Analysis}},
  year      = {2026},
  month     = feb,
  url       = {https://artificialanalysis.ai/articles/aa-wer-v2},
  note      = {Accessed: 2026-02-21}
}
